\documentclass{article}

\usepackage[final]{neurips_wrl2021}

\usepackage[utf8]{inputenc} % allow utf-8 input
\usepackage[T1]{fontenc}    % use 8-bit T1 fonts
\usepackage{hyperref}       % hyperlinks
\usepackage{url}            % simple URL typesetting
\usepackage{booktabs}       % professional-quality tables
\usepackage{amsfonts}       % blackboard math symbols
\usepackage{nicefrac}       % compact symbols for 1/2, etc.
\usepackage{microtype}      % microtypography
\usepackage{lipsum}
\usepackage{graphicx}
\graphicspath{ {./images/} }

\usepackage{pifont} % for checkmark symbols
\newcommand\cmark{\ding{51}}
\newcommand\xmark{\ding{55}}
\usepackage{caption, subcaption} % for subcaption figures
\usepackage{graphicx} % for including pdf
\usepackage{pgfplots} % for graphs
\pgfplotsset{compat=1.17}
\usepgfplotslibrary{fillbetween}
\definecolor{color1}{rgb}{0.0, 0.0, 1.0} %blue
\definecolor{color2}{rgb}{0.0, 0.8, 0.0} %green
\definecolor{color3}{rgb}{1.0, 0.0, 0.0} % red
\definecolor{color4}{rgb}{1.0, 0.7, 0.0} %pink
\definecolor{color5}{rgb}{0.0, 1.0, 1.0} %pink
\usetikzlibrary{calc}

\title{\texttt{panda-gym}: Open-source goal-conditioned environments for robotic learning}%Multi-Goal reinforcement learning environments for simulated Franka Emika Panda robot}
\author{
 Quentin Gallouédec, Nicolas Cazin, Emmanuel Dellandréa, Liming Chen\\
  École Centrale de Lyon\\
  LIRIS, CNRS UMR 5205, France\\
  \texttt{\{first.last\}@ec-lyon.fr} \\
}

\begin{document}
\maketitle
\begin{abstract}
This paper presents \texttt{panda-gym}, a set of Reinforcement Learning (RL) environments for the Franka Emika Panda robot integrated with OpenAI Gym.
Five tasks are included: reach, push, slide, pick \&  place and stack.
They all follow a Multi-Goal RL framework, allowing to use goal-oriented RL algorithms. To foster open-research, we chose to use the open-source physics engine PyBullet. 
The implementation chosen for this package allows to define very easily new tasks or new robots. 
This paper also presents a baseline of results obtained with state-of-the-art model-free off-policy
algorithms. \texttt{panda-gym} is open-source and freely available at \url{https://github.com/qgallouedec/panda-gym}.
\end{abstract}

\section{Introduction}

Recent advances in reinforcement learning applied to robotics have enabled to learn complex manipulation tasks. Nevertheless, current algorithms still struggle to solve tasks for which rewards are very sparse. Recent algorithms have contributed to the advancement in this area, but the number of interactions required to learn a satisfactory model is still very high. For the moment, learning complex tasks with sparse reward functions requires learning in simulation. A large number of physics simulator exists for various applications \citep{9386154} . For robotic manipulation, the Panda robot arm of Franka Emika is widely used. For this reason, we propose a simulated environment of this robot arm for common tasks used to evaluate RL algorithm. Contrary to what the name of the package suggests, it is possible to easily define new robots, which can be used directly with the tasks already available.
%have made it possible to meet and exceed human performance in many tasks, including board games \cite{silver2017mastering} and video games \cite{mnih2013playing}. Nevertheless, current algorithms still struggle to solve tasks for which rewards are very sparse, such as robotic environments. Recent algorithms have contributed to the advancement in this area, but the number of interactions with this type of environment to learn a satisfactory model is still very high. For the moment, learning complex tasks with sparse reward functions requires learning in simulation. 

\section{Environments}

The environments presented consist of a Panda robotic arm from Franka Emika\footnote{\url{https://www.franka.de/}}, already widely used in simulation as well as in real in many academic works. It has 7 degrees of freedom and a parallel finger gripper. The robot is simulated with the PyBullet physics engine \citep{coumans2016pybullet}, which has the advantage of being open-source and shows very good simulation performance. The environments are integrated with OpenAI Gym \citep{brockman2016openai}, allowing the use of all learning algorithms based on this API.
All the tasks presented by \citet{plappert2018multi} and \citet{andrychowicz2018hindsight} have their equivalent in this package. We have also added a stacking task, which is harder to solve than the other tasks, since two objects must be moved (instead of one for the pick \& place task). The proposed environments all follow the Multi-Goal RL framework \citep{plappert2018multi}. At each episode, a new goal is randomly generated. The type of this goal depends on the task. For example, for the \textit{reach} task, it is the position to be reached with the gripper which is randomly generated. 
The observation is therefore augmented with two additional vectors: the \textit{desired goal}, and the \textit{achieved goal}. 

\subsection{Tasks}

\begin{figure}
\centering
\begin{subfigure}[t]{0.2\textwidth}
  \centering
  \includegraphics[width=.95\linewidth]{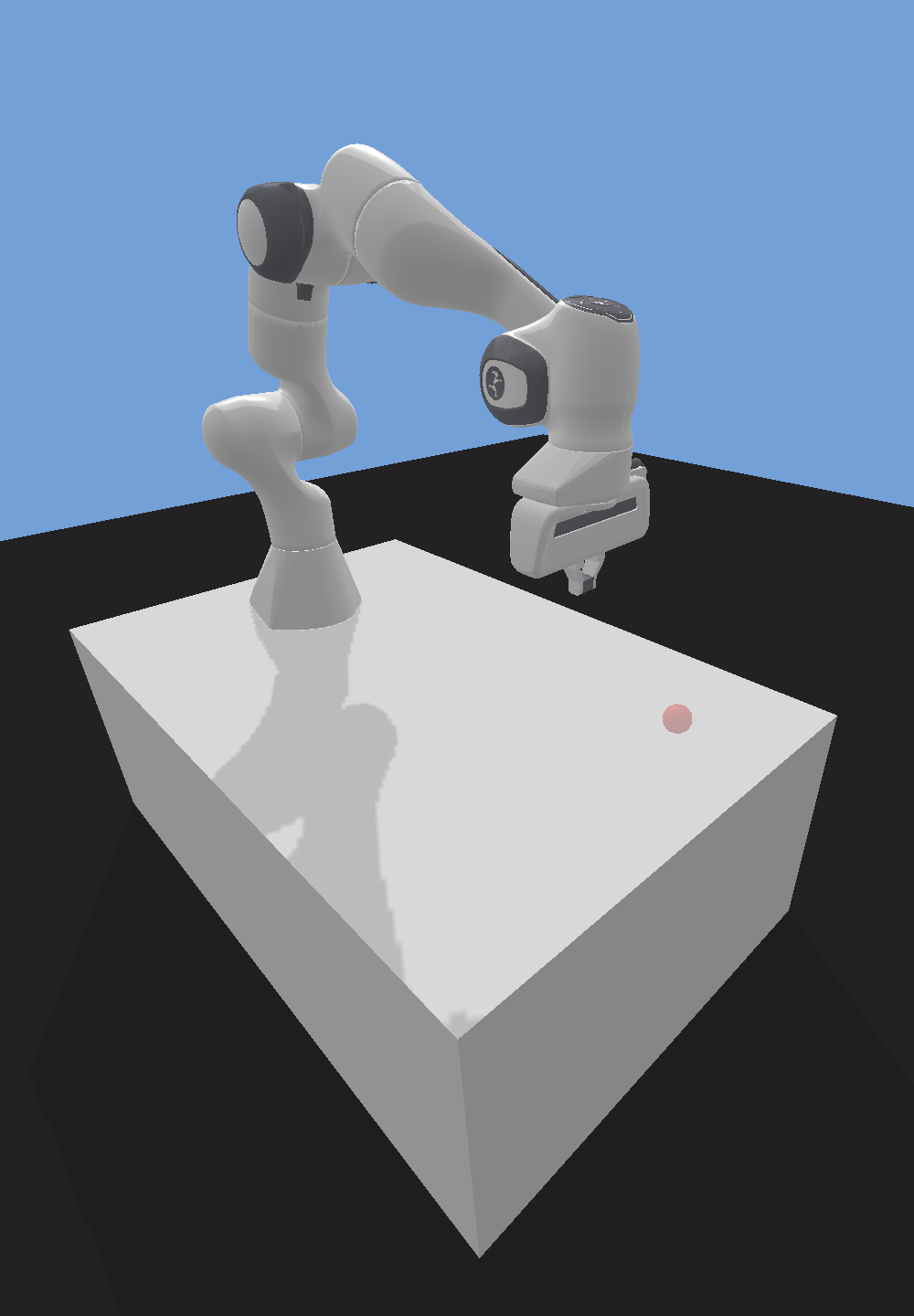}
  \caption{Reach 
  \label{fig:reach}}
\end{subfigure}%
\begin{subfigure}[t]{0.2\textwidth}
  \centering
  \includegraphics[width=.95\linewidth]{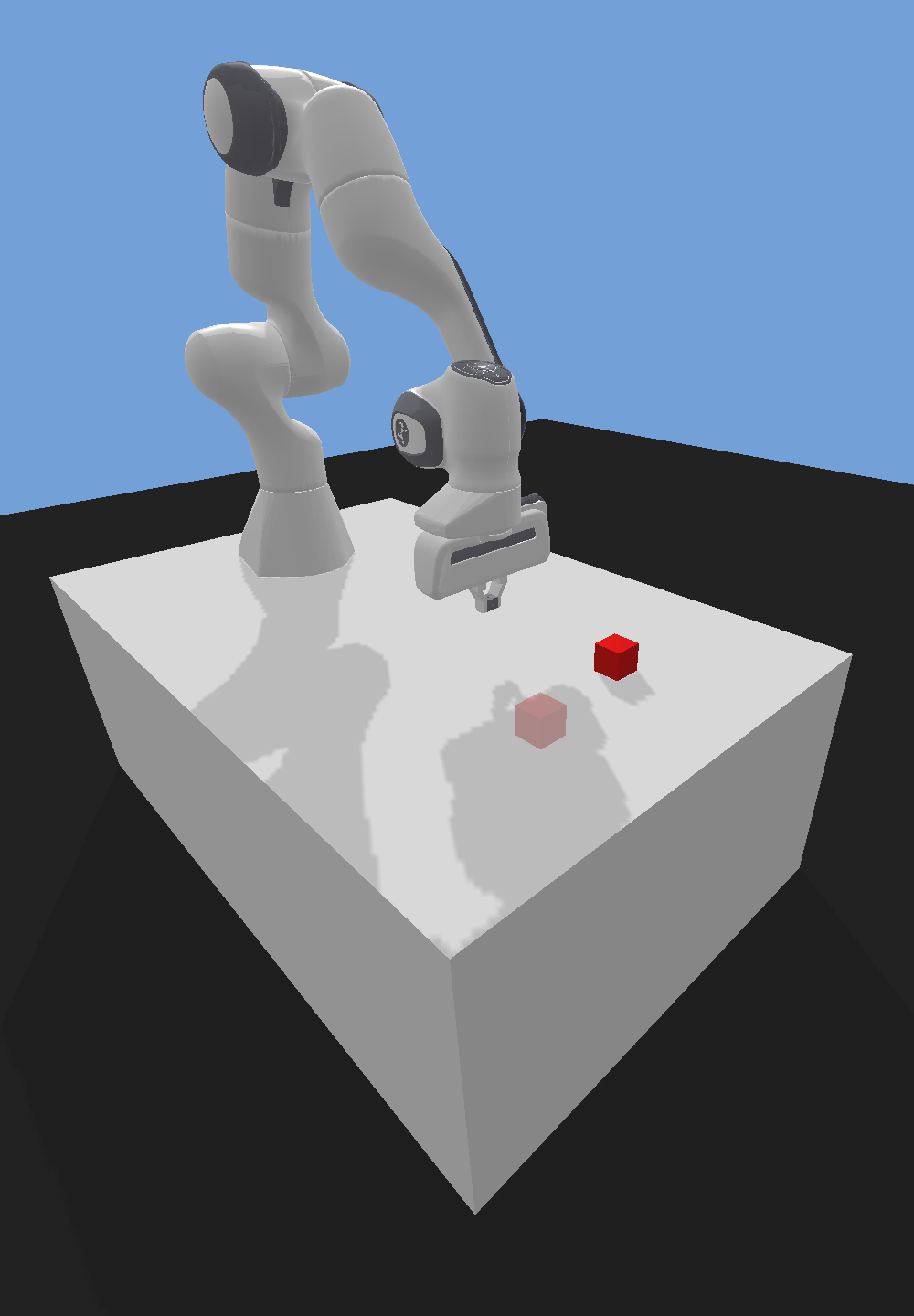}
  \caption{Push
  \label{fig:push}}
\end{subfigure}%
\begin{subfigure}[t]{0.2\textwidth}
  \centering
  \includegraphics[width=.95\linewidth]{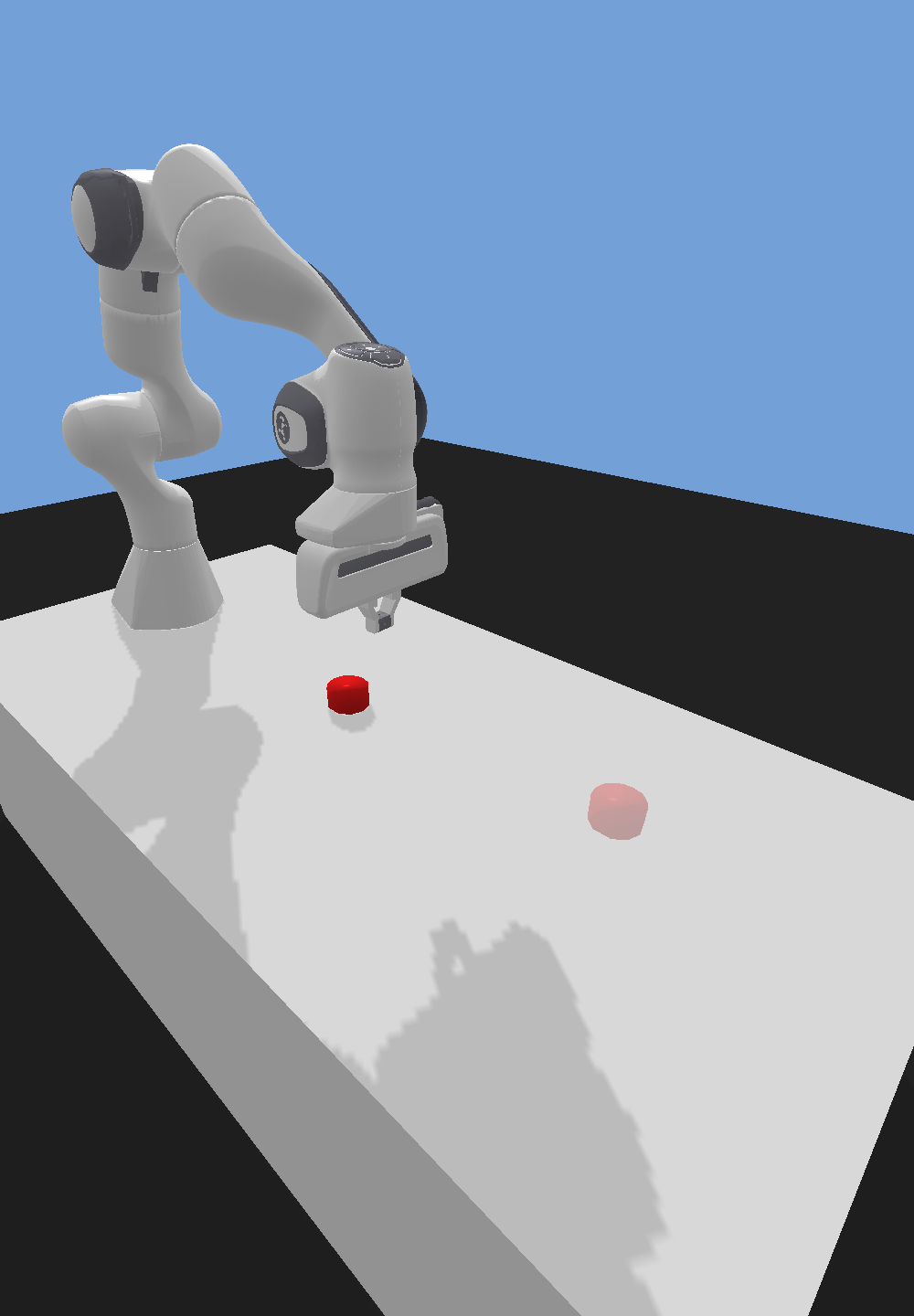}
  \caption{Slide
  \label{fig:slide}}
\end{subfigure}%
\begin{subfigure}[t]{0.2\textwidth}
  \centering
  \includegraphics[width=.95\linewidth]{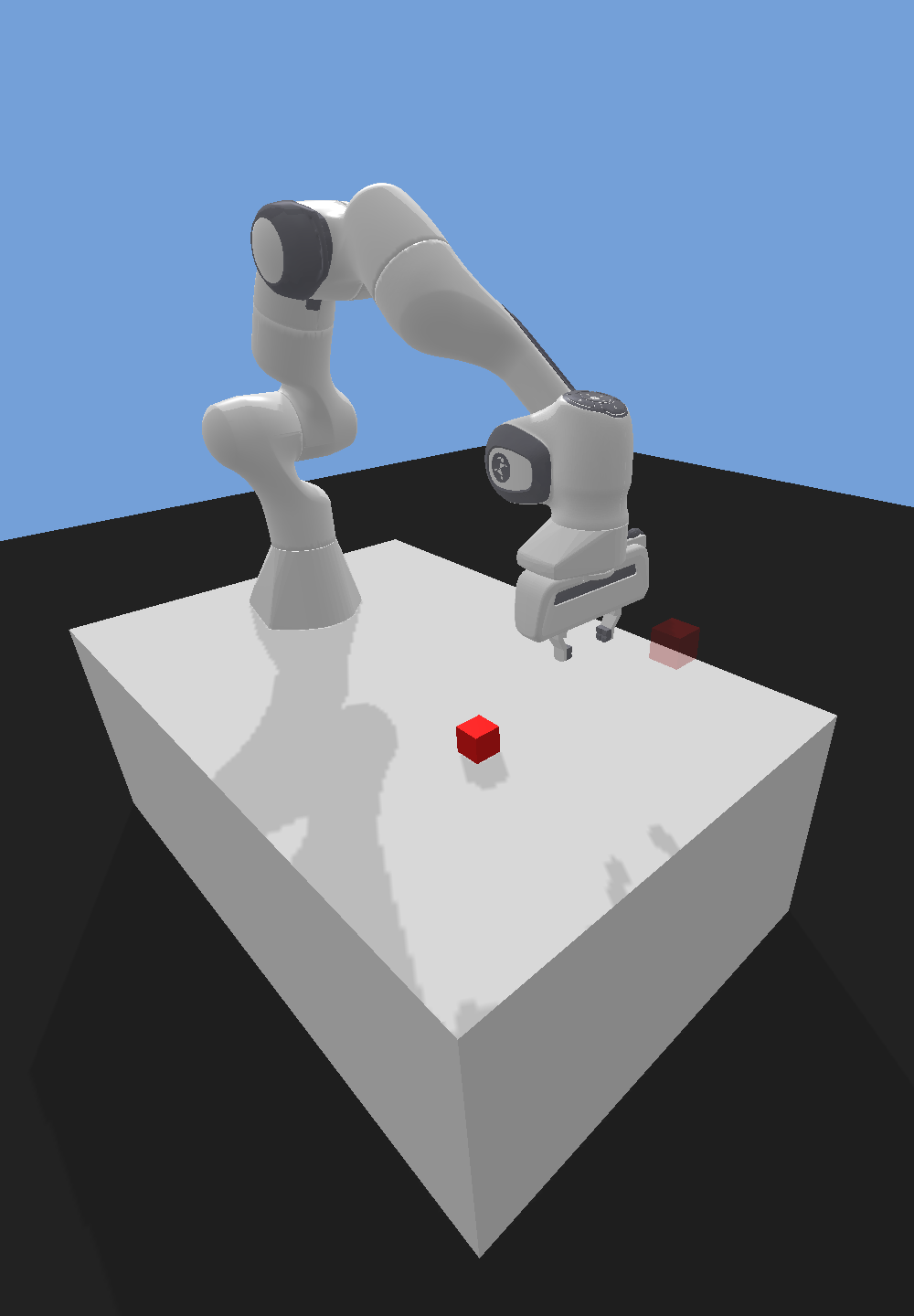}
  \caption{Pick And Place
  \label{fig:pickandplace}}
\end{subfigure}%
\begin{subfigure}[t]{0.2\textwidth}
  \centering
  \includegraphics[width=.95\linewidth]{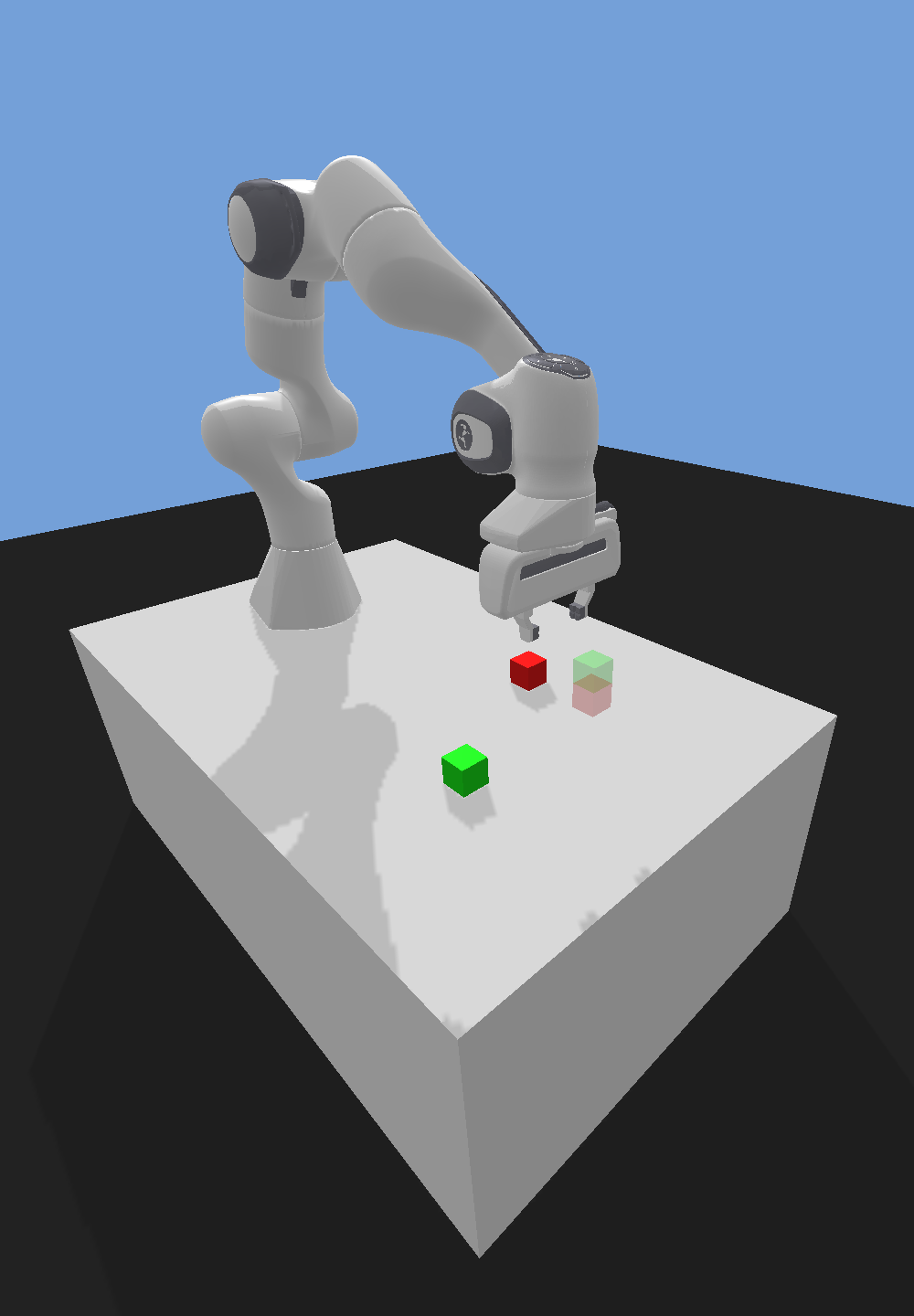}
  \caption{Stack
  \label{fig:stack}}
\end{subfigure}
\caption{Panda environments. The target positions are shaded (red and green).}
\label{fig:task_rendering}
\end{figure}

A task consists in moving either the gripper or one (or more) object(s) to a target position. A task is completed when the distance between the entity to move and the target position is less than 5 cm.
%\footnote{For the \texttt{PandaStack-v1} environment, the task is considered successful when the sum of the distances between the cube and their target position is less than 10 cm.}.
The tasks have an increasing level of difficulty. For each task, a rendering is presented in the Figure \ref{fig:task_rendering}.

\paragraph{\texttt{PandaReach-v1}}
A target position must be reached with the gripper. This target position is randomly generated in a volume of $30\textnormal{ cm} \times 30\textnormal{ cm} \times 30\textnormal{ cm}$.

\paragraph{\texttt{PandaPush-v1}}
A cube, placed on a table, must be pushed to a target position also on the table surface. The gripper is blocked closed. The target position and the initial position of the cube are randomly generated in a $30\textnormal{ cm} \times 30\textnormal{ cm}$ square around the neutral position of the robot.

\paragraph{\texttt{PandaSlide-v1}}
A flat cylinder (like an ice hockey puck) must be moved to a target position on the surface of a table. The gripper is blocked closed. 
The target position is randomly generated in a $50\textnormal{ cm} \times 50\textnormal{ cm}$ square located out of reach of the robot, in front of the neutral position. Thus, is necessary to give an impulse to the object, instead of just pushing it.

\paragraph{\texttt{PandaPickAndPlace-v1}}
A cube must be brought to a target position generated in a volume of $30\textnormal{ cm} \times 30\textnormal{ cm} \times 20\textnormal{ cm}$ above the table. 

\paragraph{\texttt{PandaStack-v1}}
Two cubes must be stacked at a target position on the table surface. The target position is generated in a square of $30\textnormal{ cm} \times 30\textnormal{ cm}$. The stacking must be done in the correct order: the red cube must be under the green cube.

\subsection{Observation and action space}

The observation space varies depending on the task. 
%Table \ref{tab:observation space} gives an overview of the composition of the observation space.
For all tasks, the observation contains the position and speed of the gripper (6 coordinates). The control of the gripper does not allow to change its orientation. Its state is therefore completely determined by these 6 coordinates.
If the task involves one or more objects, the observation space contains the position, the orientation, the linear and rotational speed (12 coordinates) for each object.
When the gripper is not constrained to be closed, the opening of the gripper (i.e. the distance between the fingers) is part of the observation space (1 coordinate).

%\begin{table}[ht]
%  \centering
%\begin{tabular}{l|c|c|c|c}
%     & Gripper & Object(s)& Gripper& Observation\\
%     & pose & pose & opening &  size\\\hline
%    \texttt{PandaReach-v1} & \cmark & \xmark & \xmark & 6\\
%    \texttt{PandaPush-v1} & \cmark & \cmark & \xmark & 18 \\
%    \texttt{PandaSlide-v1} & \cmark & \cmark & \xmark & 18 \\
%    \texttt{PandaPickAndPlace-v1} & \cmark & \cmark & \cmark & 19\\
%     \texttt{PandaStack-v1} & \cmark & \cmark & \cmark & 31
%\end{tabular}
%\caption{Components of the observation space.}
%  \label{tab:observation space}
%\end{table}

The action space is composed of the gripper movement command (3 coordinates, one for each axis of movement $x$, $y$ and $z$) and the fingers movement (1 coordinate, corresponding to the variation of the gripper opening). 
For some tasks, the gripper is blocked closed. For these tasks, the action space is only composed of the gripper motion command.

At each action of the agent, the simulator runs 20 timesteps, before giving the control back to the agent, and waiting for the next action. On the other hand, one simulator timestep represents 2 ms. The interaction frequency is thus 25 Hz. An episode is made of 50 interactions, so the duration of an episode is 2 seconds (for the stacking task, an episode lasts 100 interactions, so 4 seconds).
These durations are empirically sufficient for the realization of the corresponding tasks.

\subsection{Reward}

By default, the reward is \textit{sparse}: a reward of $0$ is obtained if the entity to move is at the desired position (with a tolerance of 5 cm), and $-1$ otherwise. 
For each environment, a variant exists in which the reward is \textit{dense}: this reward is the opposite of the distance between the entity to move and the desired position\footnote{For the stacking task, the reward is  $-\sqrt{{d_1}^2+{d_2}^2}$, where $d_i$ is the distance between the object $i$ and its desired position.}.

In general, a sparse reward function is easier to define, since it is only a question of assessing whether the task is completed in the current state.
Conversely, defining a dense reward function can be a tricky process, especially when the task implies several completion criteria.
For example, for a task consisting in moving and rotating a cube (task considered for a future version of the package, see Section \ref{sec:future_directions}), defining a dense reward function requires to assign a weight of preference to each criterion \citep{gabor1998multi, natarajan2005dynamic}. These preferences constitute additional hyperparameters.

\section{Design Decisions}

\begin{figure}[ht]
    \centering
    \includegraphics[width=0.85\textwidth]{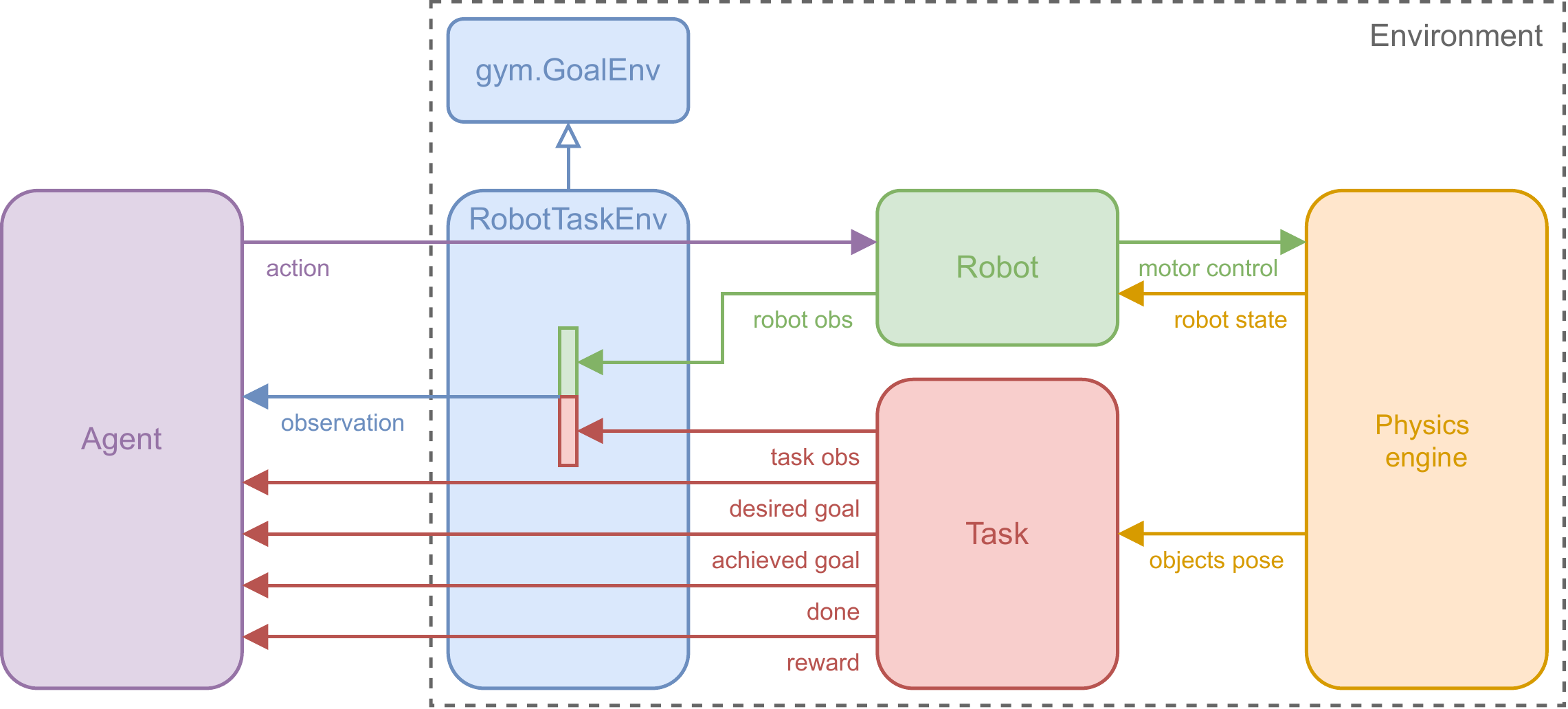}
    \caption{Code design. The task and the robot are separate, which allows them to be modular. The agent's actions are sent to the robot.}
    \label{fig:design}
\end{figure}

A robotic environment consists of a robotic arm and a task.  Conceptually, a robotic arm can perform different tasks. Similarly, a task can be performed by different robots. To allow for this flexibility, we have separated the task class from the robot class. This allows to easily define a new task without worrying about the robot that will execute it. In the same way, it is possible to define a new robot without worrying about the task to be executed. Figure \ref{fig:design} shows the chosen implementation.

The main class, called \texttt{RobotTaskEnv} contains a \texttt{robot} attribute, and a \texttt{task} attribute. When the agent takes an action,  send an action to the environment, it transfers it to the robot. The collected observation is the concatenation of the observations specific to the robot (the pose of the gripper, for example) and the observations specific to the task (the pose of the objects, for example). Finally, to follow the Multi-Goal framework, the desired goal and the achieved goal are derived from the task attribute.

The proposed environments allow fast learning, even on a computer with limited computing capacity. The PyBullet physics engine allows the parallel simulation of several scenes. Thus, the environments are compatible with learning methods that use multiple CPU cores.
Tests show that the environments are on average 9.2\% faster than their equivalents developed on MuJoCo\footnote{We measured time required to simulate $10^5$ timesteps using a single CPU core.}.

\section{Experimental results}

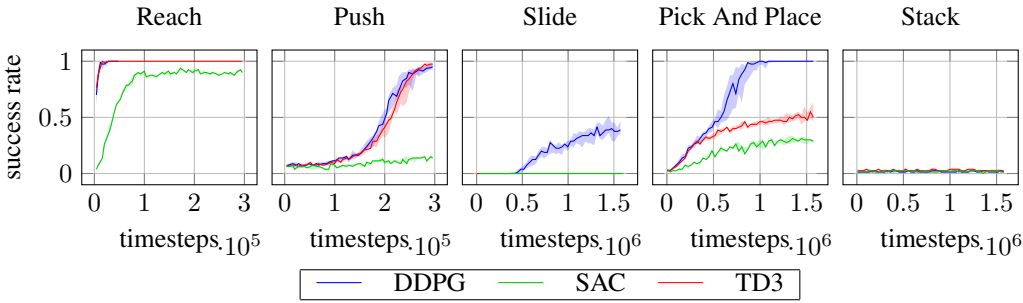
\begin{figure}[ht]
\centering
\begin{tikzpicture}
	\begin{axis}[
	    name=plot1,
	    title=Reach,
        width=0.28\textwidth,
        grid=major,
        xlabel=timesteps,
        ylabel=success rate,
        ymin=-0.1, ymax=1.1,
        ]
    \addplot [color1] table [x={timestep}, y={med}] {data/DDPG/ALL/Reach.dat}; \label{pgfplots:DDPG}
    \addplot [name path=LOWQ, draw=none, forget plot] table [x={timestep}, y={lowq}] {data/DDPG/ALL/Reach.dat};
    \addplot [name path=HIGHQ, draw=none, forget plot] table [x={timestep}, y={highq}]{data/DDPG/ALL/Reach.dat};
    \addplot [draw=none, fill=gray, opacity=0.2, color=color1, forget plot] fill between[of=LOWQ and HIGHQ];
    
    \addplot [color3] table [x={timestep}, y={med}] {data/TD3/ALL/Reach.dat};\label{pgfplots:TD3}
    \addplot [name path=LOWQ, draw=none, forget plot] table [x={timestep}, y={lowq}] {data/TD3/ALL/Reach.dat};
    \addplot [name path=HIGHQ, draw=none, forget plot] table [x={timestep}, y={highq}]{data/TD3/ALL/Reach.dat};
    \addplot [draw=none, fill=gray, opacity=0.2, color=color3, forget plot] fill between[of=LOWQ and HIGHQ];

    \addplot [color2] table [x={timestep}, y={med}] {data/SAC/ALL/Reach.dat};\label{pgfplots:SAC}
    \addplot [name path=LOWQ, draw=none, forget plot] table [x={timestep}, y={lowq}] {data/SAC/ALL/Reach.dat};
    \addplot [name path=HIGHQ, draw=none, forget plot] table [x={timestep}, y={highq}]{data/SAC/ALL/Reach.dat};
    \addplot [draw=none, fill=gray, opacity=0.2, color=color2, forget plot] fill between[of=LOWQ and HIGHQ];
    \end{axis}
    
    \begin{axis}[
	    name=plot2,
	    at={($(plot1.east)+(0.2cm,0)$)},anchor=west,
	    title=Push,
        width=0.28\textwidth,
        grid=major,
        xlabel=timesteps,
        yticklabels={,,},
        ymin=-0.1, ymax=1.1,
        % legend pos=outer north east,
        % legend cell align={left},
        ]
    \addplot [color1] table [x={timestep}, y={med}] {data/DDPG/ALL/Push.dat};
    \addplot [name path=LOWQ, draw=none, forget plot] table [x={timestep}, y={lowq}] {data/DDPG/ALL/Push.dat};
    \addplot [name path=HIGHQ, draw=none, forget plot] table [x={timestep}, y={highq}]{data/DDPG/ALL/Push.dat};
    \addplot [draw=none, fill=gray, opacity=0.2, color=color1, forget plot] fill between[of=LOWQ and HIGHQ];

    \addplot [color3] table [x={timestep}, y={med}] {data/TD3/ALL/Push.dat};
    \addplot [name path=LOWQ, draw=none, forget plot] table [x={timestep}, y={lowq}] {data/TD3/ALL/Push.dat};
    \addplot [name path=HIGHQ, draw=none, forget plot] table [x={timestep}, y={highq}]{data/TD3/ALL/Push.dat};
    \addplot [draw=none, fill=gray, opacity=0.2, color=color3, forget plot] fill between[of=LOWQ and HIGHQ];

    \addplot [color2] table [x={timestep}, y={med}] {data/SAC/ALL/Push.dat};
    \addplot [name path=LOWQ, draw=none, forget plot] table [x={timestep}, y={lowq}] {data/SAC/ALL/Push.dat};
    \addplot [name path=HIGHQ, draw=none, forget plot] table [x={timestep}, y={highq}]{data/SAC/ALL/Push.dat};
    \addplot [draw=none, fill=gray, opacity=0.2, color=color2, forget plot] fill between[of=LOWQ and HIGHQ];

    \end{axis}
    \begin{axis}[
	    name=plot3,
	    at={($(plot2.east)+(0.2cm,0)$)},anchor=west,
	    title=Slide,
        width=0.28\textwidth,
        grid=major,
        xlabel=timesteps,
        yticklabels={,,},
        ymin=-0.1, ymax=1.1,
        ]
    \addplot [color1] table [x={timestep}, y={med}] {data/DDPG/ALL/Slide.dat};
    \addplot [name path=LOWQ, draw=none, forget plot] table [x={timestep}, y={lowq}] {data/DDPG/ALL/Slide.dat};
    \addplot [name path=HIGHQ, draw=none, forget plot] table [x={timestep}, y={highq}]{data/DDPG/ALL/Slide.dat};
    \addplot [draw=none, fill=gray, opacity=0.2, color=color1, forget plot] fill between[of=LOWQ and HIGHQ];

    \addplot [color3] table [x={timestep}, y={med}] {data/TD3/ALL/Slide.dat};
    \addplot [name path=LOWQ, draw=none, forget plot] table [x={timestep}, y={lowq}] {data/TD3/ALL/Slide.dat};
    \addplot [name path=HIGHQ, draw=none, forget plot] table [x={timestep}, y={highq}]{data/TD3/ALL/Slide.dat};
    \addplot [draw=none, fill=gray, opacity=0.2, color=color3, forget plot] fill between[of=LOWQ and HIGHQ];

    \addplot [color2] table [x={timestep}, y={med}] {data/SAC/ALL/Slide.dat};
    \addplot [name path=LOWQ, draw=none, forget plot] table [x={timestep}, y={lowq}] {data/SAC/ALL/Slide.dat};
    \addplot [name path=HIGHQ, draw=none, forget plot] table [x={timestep}, y={highq}]{data/SAC/ALL/Slide.dat};
    \addplot [draw=none, fill=gray, opacity=0.2, color=color2, forget plot] fill between[of=LOWQ and HIGHQ];

    \end{axis}
    
    \begin{axis}[
	    name=plot4,
	    at={($(plot3.east)+(0.2cm,0)$)},anchor=west,
	    title=Pick And Place,
        width=0.28\textwidth,
        grid=major,
        xlabel=timesteps,
        yticklabels={,,},
        ymin=-0.1, ymax=1.1,
        ]
    \addplot [color1] table [x={timestep}, y={med}] {data/DDPG/ALL/PickAndPlace.dat};
    \addplot [name path=LOWQ, draw=none, forget plot] table [x={timestep}, y={lowq}] {data/DDPG/ALL/PickAndPlace.dat};
    \addplot [name path=HIGHQ, draw=none, forget plot] table [x={timestep}, y={highq}]{data/DDPG/ALL/PickAndPlace.dat};
    \addplot [draw=none, fill=gray, opacity=0.2, color=color1, forget plot] fill between[of=LOWQ and HIGHQ];

    \addplot [color3] table [x={timestep}, y={med}] {data/TD3/ALL/PickAndPlace.dat};
    \addplot [name path=LOWQ, draw=none, forget plot] table [x={timestep}, y={lowq}] {data/TD3/ALL/PickAndPlace.dat};
    \addplot [name path=HIGHQ, draw=none, forget plot] table [x={timestep}, y={highq}]{data/TD3/ALL/PickAndPlace.dat};
    \addplot [draw=none, fill=gray, opacity=0.2, color=color3, forget plot] fill between[of=LOWQ and HIGHQ];

    \addplot [color2] table [x={timestep}, y={med}] {data/SAC/ALL/PickAndPlace.dat};
    \addplot [name path=LOWQ, draw=none, forget plot] table [x={timestep}, y={lowq}] {data/SAC/ALL/PickAndPlace.dat};
    \addplot [name path=HIGHQ, draw=none, forget plot] table [x={timestep}, y={highq}]{data/SAC/ALL/PickAndPlace.dat};
    \addplot [draw=none, fill=gray, opacity=0.2, color=color2, forget plot] fill between[of=LOWQ and HIGHQ];

    \end{axis}
    
    \begin{axis}[
	    name=plot5,
	    at={($(plot4.east)+(0.2cm,0)$)},anchor=west,
	    title=Stack,
        width=0.28\textwidth,
        grid=major,
        xlabel=timesteps,
        yticklabels={,,},
        ymin=-0.1, ymax=1.1,
        ]
    \addplot [color1] table [x={timestep}, y={med}] {data/DDPG/ALL/Stack.dat};
    \addplot [name path=LOWQ, draw=none, forget plot] table [x={timestep}, y={lowq}] {data/DDPG/ALL/Stack.dat};
    \addplot [name path=HIGHQ, draw=none, forget plot] table [x={timestep}, y={highq}]{data/DDPG/ALL/Stack.dat};
    \addplot [draw=none, fill=gray, opacity=0.2, color=color1, forget plot] fill between[of=LOWQ and HIGHQ];

    \addplot [color3] table [x={timestep}, y={med}] {data/TD3/ALL/Stack.dat};
    \addplot [name path=LOWQ, draw=none, forget plot] table [x={timestep}, y={lowq}] {data/TD3/ALL/Stack.dat};
    \addplot [name path=HIGHQ, draw=none, forget plot] table [x={timestep}, y={highq}]{data/TD3/ALL/Stack.dat};
    \addplot [draw=none, fill=gray, opacity=0.2, color=color3, forget plot] fill between[of=LOWQ and HIGHQ];
    
    \addplot [color2] table [x={timestep}, y={med}] {data/SAC/ALL/Stack.dat};
    \addplot [name path=LOWQ, draw=none, forget plot] table [x={timestep}, y={lowq}] {data/SAC/ALL/Stack.dat};
    \addplot [name path=HIGHQ, draw=none, forget plot] table [x={timestep}, y={highq}]{data/SAC/ALL/Stack.dat};
    \addplot [draw=none, fill=gray, opacity=0.2, color=color2, forget plot] fill between[of=LOWQ and HIGHQ];

    \end{axis}
    \draw (0:0) node[at={($(plot3.south)+(0,-1.0cm)$)}, anchor=north]{
    \begin{tabular}{|clclcl|}
    \hline
    \ref{pgfplots:DDPG} & DDPG &
    \ref{pgfplots:SAC} & SAC &
    \ref{pgfplots:TD3} & TD3
    \\ \hline
    \end{tabular}};
    
\end{tikzpicture}
\caption{Success rates for the five Panda environments. We repeat each experiment with 21 different random seeds. Median rates are solid lines and interquartile range are shaded areas. We represent the results for the DDPG, SAC and TD3 algorithms, all three ran with HER. The horizontal axis corresponds to the total number of interaction with the environment.}
\label{fig:learning_curves}
\end{figure}

% \subsection{Learning with HER}
The length of a trajectory is 50 timesteps, except for the stacking task, for which we chose a length of 100 timesteps, due to its higher complexity. At the end of each trajectory, the environment is reset and a new goal is randomly generated. The learning has been distributed on 8 CPU cores. Each core generates trajectories and all these trajectories are stored in a common replay buffer. The results are the success rate evaluated over 80 test episodes, regularly over the course of learning.
We give a baseline of the results we obtain for three off-policy algorithms from the recent literature used with Hindsight Experience Replay (HER) \citep{andrychowicz2018hindsight}: Deep Deterministic Policy Gradient (DDPG) \citep{lillicrap2015continuous}, Soft Actor-Critic (SAC) \citep{haarnoja2018soft} and Twin Delayed DDPG (TD3) \citep{fujimoto2018addressing}.
The implementation of DDPG used for the training is the one proposed by \citet{baselines}. The appropriate modifications have been made to DDPG to implement TD3 and SAC\footnote{\url{https://github.com/qgallouedec/baselines}}. The hyperparameters are available in Appendix \ref{hyperparameters}. The learning curves are shown in Figure \ref{fig:learning_curves}. Note that the horizontal axis corresponds to the total number of interaction with the environment. The learning curves are therefore independent of the number of workers used to collect these interactions.

%As the complexity of the tasks increases, the number of timesteps needed to solve them increases. 

The number of timesteps needed to resolve a task depend on the task and the algorithm. For DDPG, the success rate reaches 100\% for the reach and push tasks, after $10^4$ and $3\times10^4$ timesteps, respectively. It reaches about 50\% for the slide and pick and place tasks, after $6\times 10^5$ and $1.6\times10^6$ timesteps, respectively. 
The success rate for the stacking task remains close to 0 after $1.6\times10^6$ timesteps of training. The presented algorithms do not allow to solve it in this amount of timesteps.

We notice that for TD3 and SAC, the ablation of the clipped double-$Q$ trick allows a significant increase of the results in multiple environments. The set of curves representing the results of the distinct ablations is given in Appendix \ref{appendix:Ablation}. Appendix \ref{appendix:policies} shows an overview of the policies at the end of the training for the four task that are solved or partially solved.

% \subsection{Simulation performance (not sure about the utility of this section)}

% The proposed environments allow a fast learning process even on a computer with limited computing capacity. A comparison of the simulation times is proposed in table \ref{table:performances}. These tests were performed on the same laptop.

% \begin{table}[ht]
%   \centering
% \begin{tabular}{l|c|c|c}
%      & Panda and  & Fetch and \\
%       & PyBullet & MuJoCo \\ \hline
%     Reach & 18.06 s & 19.31 s & 6.9\%\\
%     Push & 20.66 s & 22.91 s & 10.9\%\\
%     Slide & 22.51 s & 23.35 s & 3.7\%\\
%     Pick and Place & 18.99 s & 21.96 s & 15.6\%\\
% \end{tabular}
% \caption{Components of the observation space}
%   \label{tab:performances}
% \end{table}

\section{Conclusion and future works}
\label{sec:future_directions}
In this paper, we have described \texttt{panda-gym}, a free and open-source package which allows to define robotic tasks, 5 of which are present in the current version.
They allow to evaluate the reinforcement learning algorithms in the context of complex robotic tasks.
The architectural choices allow to define very easily new tasks and new robots. Finally, the state of the art algorithms allow to solve some tasks, while others remain unsolved.

We are planning to add to the presented tasks some new and very used tasks such as the peg-in-hole insertion \citep{lee2019making} or the cube flipping \citep{robogym2020}. On the other hand, we also plan to give the possibility to control the robot directly with joint values. This would require the agent to learn by itself the inverse robot dynamics. Finally, to better fit with reality, we plan to add the possibility to use an observation space that would include several modalities, such as a RGB camera, a depth camera, a force sensor or a tactile sensor.

\newpage
% \section*{Acknowledgment}

% This work was supported in part by the EU FEDER funding through the FUI PIKAFLEX project, by the French National Research Agency, \textit{l'Agence Nationale de Recherche} (ANR) through the ARES LabCom under grant ANR 16-LCV2-0012-01, the LEARN-REAL project within the the EU CHIST-ERA program under grant ANR-18-CHR3-0002-01 and Chiron project within the trilateral France-Germany-Japan program on AI.

% This work was performed using HPC resources from GENCI-IDRIS (Grant 20XX-[AD011012172]).

\bibliographystyle{unsrtnat}  
\bibliography{references}

\appendix
\newpage
\section{Hyperparameters}
\label{hyperparameters}

All experiments in this paper use the hyperparameters presented in Table \ref{tab:hyperparam}.  They have been chosen identical to those submitted by \citet{plappert2018multi}. 

\begin{table}[ht]
    \centering
    \begin{tabular}{lll}
    \toprule
    Actor and Critic & Network type & Multi-layer perceptron \\
        & Network size & 3 layers of 256 nodes \\
        & Optimizer &  Adam \citep{kingma2014adam} \\
        & Learning rate & $0.001$\\
        & Polyak-averaging \citeyearpar{polyak1992acceleration} & $0.95$\\
        & L2 normalisation coefficient & $1.0$\\ \hline
    Observation & Clipping & $[-200, 200]$\\\hline
    Action & Clipping & $[-1, 1]$\\
        & Probability of random action& $0.3$  (DDPG and TD3 only) \\
        & Scale of additive gaussian noise& $0.2$  (DDPG and TD3 only) \\
        & Number of HER per transition ($k$)& $4$\\ \hline
    Training & Episode length & $50$ ($100$ for \texttt{PandaStack-v1})\\
        & Testing & Every 80 episodes\\
        & Number of testing episodes & $80$\\
        & Replay buffer size & $10^6$ transitions\\
        & Batch size & $256$ \\
        & Policy delay & $2$  (TD3 only)\\
        & Policy noise & $0.2$  (TD3 only)\\
        & Policy noise clip & $[-0.5, 0.5]$  (TD3 only)\\
        & $\alpha$ & $0.2$  (SAC only)\\
    \bottomrule
    \end{tabular}
    \caption{Hyperparameters used for the experiments}
    \label{tab:hyperparam}
\end{table}

\newpage
\section{Full results and ablations study}
\label{appendix:Ablation}

\paragraph{Ablating the clipped double-$Q$ trick}

In the TD3 algorithm, a transition $(s, a, r, s', d)$ is sampled from the replay buffer, where $s$ is a state, $a$ the action, $r$ the reward, $s'$ the next state, and $d$ a boolean value indicating if $s'$ is terminal. The target value $y$ is computed as follows:

\begin{equation}
    y(r, s', d)=r+\gamma (1-d)\min_{i=1,2}Q_{\phi_{\mathrm{targ},i}}(s', a'(s'))
\end{equation}

where $\gamma$ is the discount factor, $Q_{\phi_{\mathrm{targ},1}}$ and $Q_{\phi_{\mathrm{targ},2}}$ are the target $Q$-networks, and $a'(s')$ is the target action, resulting from the target policy smoothing. The ablation of the clipped double-$Q$ trick in TD3 consists in replacing the target value by

\begin{equation}
    y(r, s', d)=r+\gamma (1-d)Q_{\phi_{\mathrm{targ},1}}(s', a'(s'))
\end{equation}

In the SAC algorithm, the target value $y$ is computed as follow:

\begin{equation}
    y(r, s', d)=r+\gamma (1-d)\left(\min_{i=1,2}Q_{\phi_{\mathrm{targ},i}}(s', \tilde{a}') - \alpha \log \pi_\theta (\tilde{a}'\mid s')\right), \quad \tilde{a}'\sim \pi_\theta(\cdot\mid s')
\end{equation}

where $\alpha$ is the entropy regularization coefficient and $\pi_\theta$ the stochastic policy of the actor.
The ablation of the clipped double-$Q$ trick in SAC consists in replacing the target value by

\begin{equation}
    y(r, s', d)=r+\gamma (1-d)\left(Q_{\phi_{\mathrm{targ},1}}(s', \tilde{a}') - \alpha \log \pi_\theta (\tilde{a}'\mid s')\right), \quad \tilde{a}'\sim \pi_\theta(\cdot\mid s')
\end{equation}

Once the clipped double-$Q$ trick is ablated, the second $Q$-network (denoted $Q_{\phi_{\mathrm{targ},2}}$) no longer exists.

\paragraph{Ablating HER} After each episode, the agent stores in the replay buffer each transition with the initial goal. For each stored transition, it stores $k$ identical transitions, with a different goal, corresponding to a goal achieved later, during the episode. The ablation of HER consists in removing this part of the algorithm.

\begin{figure}
\centering
\begin{tikzpicture}
	\begin{axis}[
	    name=ReachDDPG,
	    title=DDPG,
        width=0.3\textwidth,
        grid=major,
        scaled x ticks = false,
        xticklabels={,,},
        ylabel=success rate,
        ymin=-0.1, ymax=1.1,
        ]
        \addplot [color1] table [x={timestep}, y={med}] {data/DDPG/ALL/Reach.dat};
        \addplot [name path=LOWQ, draw=none, forget plot] table [x={timestep}, y={lowq}] {data/DDPG/ALL/Reach.dat};
        \addplot [name path=HIGHQ, draw=none, forget plot] table [x={timestep}, y={highq}]{data/DDPG/ALL/Reach.dat};
        \addplot [draw=none, fill=gray, opacity=0.2, color=color1, forget plot] fill between[of=LOWQ and HIGHQ];
        
        \addplot [color2] table [x={timestep}, y={med}] {data/DDPG/no_HER/Reach.dat};
        \addplot [name path=LOWQ, draw=none, forget plot] table [x={timestep}, y={lowq}] {data/DDPG/no_HER/Reach.dat};
        \addplot [name path=HIGHQ, draw=none, forget plot] table [x={timestep}, y={highq}]{data/DDPG/no_HER/Reach.dat};
        \addplot [draw=none, fill=gray, opacity=0.2, color=color2, forget plot] fill between[of=LOWQ and HIGHQ];
    \end{axis}
    
    \draw (0:0) node[at={($(ReachDDPG.west)+(-1.5cm,0)$)},  rotate=90]{\footnotesize\texttt{PandaReach-v1}};

    \begin{axis}[
	    name=ReachTD3,
	    at={($(ReachDDPG.east)+(0.5cm,0)$)},anchor=west,
	    title=TD3,
        width=0.3\textwidth,
        grid=major,
        scaled x ticks = false,
        xticklabels={,,},
        yticklabels={,,},
        ymin=-0.1, ymax=1.1,
        ]
        \addplot [color1] table [x={timestep}, y={med}] {data/TD3/ALL/Reach.dat};
        \addplot [name path=LOWQ, draw=none, forget plot] table [x={timestep}, y={lowq}] {data/TD3/ALL/Reach.dat};
        \addplot [name path=HIGHQ, draw=none, forget plot] table [x={timestep}, y={highq}]{data/TD3/ALL/Reach.dat};
        \addplot [draw=none, fill=gray, opacity=0.2, color=color1, forget plot] fill between[of=LOWQ and HIGHQ];
        
        \addplot [color2] table [x={timestep}, y={med}] {data/TD3/no_HER/Reach.dat};
        \addplot [name path=LOWQ, draw=none, forget plot] table [x={timestep}, y={lowq}] {data/TD3/no_HER/Reach.dat};
        \addplot [name path=HIGHQ, draw=none, forget plot] table [x={timestep}, y={highq}]{data/TD3/no_HER/Reach.dat};
        \addplot [draw=none, fill=gray, opacity=0.2, color=color2, forget plot] fill between[of=LOWQ and HIGHQ];
        
        \addplot [color3] table [x={timestep}, y={med}] {data/TD3/no_DQT/Reach.dat};
        \addplot [name path=LOWQ, draw=none, forget plot] table [x={timestep}, y={lowq}] {data/TD3/no_DQT/Reach.dat};
        \addplot [name path=HIGHQ, draw=none, forget plot] table [x={timestep}, y={highq}]{data/TD3/no_DQT/Reach.dat};
        \addplot [draw=none, fill=gray, opacity=0.2, color=color3, forget plot] fill between[of=LOWQ and HIGHQ];
    \end{axis}

    \begin{axis}[
	    name=ReachSAC,
	    at={($(ReachTD3.east)+(0.5cm,0)$)},anchor=west,
	    title=SAC,
        width=0.3\textwidth,
        grid=major,
        scaled x ticks = false,
        xticklabels={,,},
        yticklabels={,,},
        ymin=-0.1, ymax=1.1,
        ]
        \addplot [color1] table [x={timestep}, y={med}] {data/SAC/ALL/Reach.dat};%\label{plot:no_ablation}
        \addplot [name path=LOWQ, draw=none, forget plot] table [x={timestep}, y={lowq}] {data/SAC/ALL/Reach.dat};
        \addplot [name path=HIGHQ, draw=none, forget plot] table [x={timestep}, y={highq}]{data/SAC/ALL/Reach.dat};
        \addplot [draw=none, fill=gray, opacity=0.2, color=color1, forget plot] fill between[of=LOWQ and HIGHQ];
        
        \addplot [color2] table [x={timestep}, y={med}] {data/SAC/no_HER/Reach.dat};%\label{plot:no_HER}
        \addplot [name path=LOWQ, draw=none, forget plot] table [x={timestep}, y={lowq}] {data/SAC/no_HER/Reach.dat};
        \addplot [name path=HIGHQ, draw=none, forget plot] table [x={timestep}, y={highq}]{data/SAC/no_HER/Reach.dat};
        \addplot [draw=none, fill=gray, opacity=0.2, color=color2, forget plot] fill between[of=LOWQ and HIGHQ];
        
        \addplot [color3] table [x={timestep}, y={med}] {data/SAC/no_DQT/Reach.dat};%\label{plot:no_DQT}
        \addplot [name path=LOWQ, draw=none, forget plot] table [x={timestep}, y={lowq}] {data/SAC/no_DQT/Reach.dat};
        \addplot [name path=HIGHQ, draw=none, forget plot] table [x={timestep}, y={highq}]{data/SAC/no_DQT/Reach.dat};
        \addplot [draw=none, fill=gray, opacity=0.2, color=color3, forget plot] fill between[of=LOWQ and HIGHQ];
    \end{axis}

    \begin{axis}[
	    name=PushDDPG,
	    at={($(ReachDDPG.south)+(0, -1.0cm)$)},anchor=north,
        width=0.3\textwidth,
        grid=major,
        ylabel=success rate,
        xlabel=timesteps,
        ymin=-0.1, ymax=1.1,
        ]
        \addplot [color1] table [x={timestep}, y={med}] {data/DDPG/ALL/Push.dat};
        \addplot [name path=LOWQ, draw=none, forget plot] table [x={timestep}, y={lowq}] {data/DDPG/ALL/Push.dat};
        \addplot [name path=HIGHQ, draw=none, forget plot] table [x={timestep}, y={highq}]{data/DDPG/ALL/Push.dat};
        \addplot [draw=none, fill=gray, opacity=0.2, color=color1, forget plot] fill between[of=LOWQ and HIGHQ];
        
        \addplot [color2] table [x={timestep}, y={med}] {data/DDPG/no_HER/Push.dat};
        \addplot [name path=LOWQ, draw=none, forget plot] table [x={timestep}, y={lowq}] {data/DDPG/no_HER/Push.dat};
        \addplot [name path=HIGHQ, draw=none, forget plot] table [x={timestep}, y={highq}]{data/DDPG/no_HER/Push.dat};
        \addplot [draw=none, fill=gray, opacity=0.2, color=color2, forget plot] fill between[of=LOWQ and HIGHQ];
    \end{axis}
    
    \draw (0:0) node[at={($(PushDDPG.west)+(-1.5cm,0)$)},  rotate=90]{\footnotesize\texttt{PandaPush-v1}};
    
    \begin{axis}[
	    name=PushTD3,
	    at={($(PushDDPG.east)+(0.5cm,0)$)}, anchor=west,
        width=0.3\textwidth,
        grid=major,
        yticklabels={,,},
        xlabel=timesteps,
        ymin=-0.1, ymax=1.1,
        ]
        \addplot [color1] table [x={timestep}, y={med}] {data/TD3/ALL/Push.dat};
        \addplot [name path=LOWQ, draw=none, forget plot] table [x={timestep}, y={lowq}] {data/TD3/ALL/Push.dat};
        \addplot [name path=HIGHQ, draw=none, forget plot] table [x={timestep}, y={highq}]{data/TD3/ALL/Push.dat};
        \addplot [draw=none, fill=gray, opacity=0.2, color=color1, forget plot] fill between[of=LOWQ and HIGHQ];
        
        \addplot [color2] table [x={timestep}, y={med}] {data/TD3/no_HER/Push.dat};
        \addplot [name path=LOWQ, draw=none, forget plot] table [x={timestep}, y={lowq}] {data/TD3/no_HER/Push.dat};
        \addplot [name path=HIGHQ, draw=none, forget plot] table [x={timestep}, y={highq}]{data/TD3/no_HER/Push.dat};
        \addplot [draw=none, fill=gray, opacity=0.2, color=color2, forget plot] fill between[of=LOWQ and HIGHQ];
        
        \addplot [color3] table [x={timestep}, y={med}] {data/TD3/no_DQT/Push.dat};
        \addplot [name path=LOWQ, draw=none, forget plot] table [x={timestep}, y={lowq}] {data/TD3/no_DQT/Push.dat};
        \addplot [name path=HIGHQ, draw=none, forget plot] table [x={timestep}, y={highq}]{data/TD3/no_DQT/Push.dat};
        \addplot [draw=none, fill=gray, opacity=0.2, color=color3, forget plot] fill between[of=LOWQ and HIGHQ];
    \end{axis}
    
    \begin{axis}[
	    name=PushSAC,
	    at={($(PushTD3.east)+(0.5cm,0)$)},anchor=west,
	    width=0.3\textwidth,
        grid=major,
        yticklabels={,,},
        ymin=-0.1, ymax=1.1,
        xlabel=timesteps,
        ]
        \addplot [color1] table [x={timestep}, y={med}] {data/SAC/ALL/Push.dat};
        \addplot [name path=LOWQ, draw=none, forget plot] table [x={timestep}, y={lowq}] {data/SAC/ALL/Push.dat};
        \addplot [name path=HIGHQ, draw=none, forget plot] table [x={timestep}, y={highq}]{data/SAC/ALL/Push.dat};
        \addplot [draw=none, fill=gray, opacity=0.2, color=color1, forget plot] fill between[of=LOWQ and HIGHQ];
        
        \addplot [color2] table [x={timestep}, y={med}] {data/SAC/no_HER/Push.dat};
        \addplot [name path=LOWQ, draw=none, forget plot] table [x={timestep}, y={lowq}] {data/SAC/no_HER/Push.dat};
        \addplot [name path=HIGHQ, draw=none, forget plot] table [x={timestep}, y={highq}]{data/SAC/no_HER/Push.dat};
        \addplot [draw=none, fill=gray, opacity=0.2, color=color2, forget plot] fill between[of=LOWQ and HIGHQ];
        
        \addplot [color3] table [x={timestep}, y={med}] {data/SAC/no_DQT/Push.dat};
        \addplot [name path=LOWQ, draw=none, forget plot] table [x={timestep}, y={lowq}] {data/SAC/no_DQT/Push.dat};
        \addplot [name path=HIGHQ, draw=none, forget plot] table [x={timestep}, y={highq}]{data/SAC/no_DQT/Push.dat};
        \addplot [draw=none, fill=gray, opacity=0.2, color=color3, forget plot] fill between[of=LOWQ and HIGHQ];
    \end{axis}
    
    \begin{axis}[
	    name=SlideDDPG,
	    at={($(PushDDPG.south)+(0, -2.0cm)$)},anchor=north,
        width=0.3\textwidth,
        grid=major,
        scaled x ticks = false,
        ylabel=success rate,
        xticklabels={,,},
        ymin=-0.1, ymax=1.1,
        ]
        \addplot [color1] table [x={timestep}, y={med}] {data/DDPG/ALL/Slide.dat};
        \addplot [name path=LOWQ, draw=none, forget plot] table [x={timestep}, y={lowq}] {data/DDPG/ALL/Slide.dat};
        \addplot [name path=HIGHQ, draw=none, forget plot] table [x={timestep}, y={highq}]{data/DDPG/ALL/Slide.dat};
        \addplot [draw=none, fill=gray, opacity=0.2, color=color1, forget plot] fill between[of=LOWQ and HIGHQ];
        
        \addplot [color2] table [x={timestep}, y={med}] {data/DDPG/no_HER/Slide.dat};
        \addplot [name path=LOWQ, draw=none, forget plot] table [x={timestep}, y={lowq}] {data/DDPG/no_HER/Slide.dat};
        \addplot [name path=HIGHQ, draw=none, forget plot] table [x={timestep}, y={highq}]{data/DDPG/no_HER/Slide.dat};
        \addplot [draw=none, fill=gray, opacity=0.2, color=color2, forget plot] fill between[of=LOWQ and HIGHQ];
    \end{axis}
    
    \draw (0:0) node[at={($(SlideDDPG.west)+(-1.5cm,0)$)},  rotate=90]{\footnotesize\texttt{PandaSlide-v1}};
    
    \begin{axis}[
	    name=SlideTD3,
	    at={($(SlideDDPG.east)+(0.5cm,0)$)}, anchor=west,
        width=0.3\textwidth,
        grid=major,
        scaled x ticks = false,
        yticklabels={,,},
        xticklabels={,,},
        ymin=-0.1, ymax=1.1,
        ]
        \addplot [color1] table [x={timestep}, y={med}] {data/TD3/ALL/Slide.dat};
        \addplot [name path=LOWQ, draw=none, forget plot] table [x={timestep}, y={lowq}] {data/TD3/ALL/Slide.dat};
        \addplot [name path=HIGHQ, draw=none, forget plot] table [x={timestep}, y={highq}]{data/TD3/ALL/Slide.dat};
        \addplot [draw=none, fill=gray, opacity=0.2, color=color1, forget plot] fill between[of=LOWQ and HIGHQ];
        
        \addplot [color2] table [x={timestep}, y={med}] {data/TD3/no_HER/Slide.dat};
        \addplot [name path=LOWQ, draw=none, forget plot] table [x={timestep}, y={lowq}] {data/TD3/no_HER/Slide.dat};
        \addplot [name path=HIGHQ, draw=none, forget plot] table [x={timestep}, y={highq}]{data/TD3/no_HER/Slide.dat};
        \addplot [draw=none, fill=gray, opacity=0.2, color=color2, forget plot] fill between[of=LOWQ and HIGHQ];
        
        \addplot [color3] table [x={timestep}, y={med}] {data/TD3/no_DQT/Slide.dat};
        \addplot [name path=LOWQ, draw=none, forget plot] table [x={timestep}, y={lowq}] {data/TD3/no_DQT/Slide.dat};
        \addplot [name path=HIGHQ, draw=none, forget plot] table [x={timestep}, y={highq}]{data/TD3/no_DQT/Slide.dat};
        \addplot [draw=none, fill=gray, opacity=0.2, color=color3, forget plot] fill between[of=LOWQ and HIGHQ];
    \end{axis}
    
    \begin{axis}[
	    name=SlideSAC,
	    at={($(SlideTD3.east)+(0.5cm,0)$)},anchor=west,
	    width=0.3\textwidth,
        grid=major,
        scaled x ticks = false,
        yticklabels={,,},
        xticklabels={,,},
        ymin=-0.1, ymax=1.1,
        ]
        \addplot [color1] table [x={timestep}, y={med}] {data/SAC/ALL/Slide.dat};
        \addplot [name path=LOWQ, draw=none, forget plot] table [x={timestep}, y={lowq}] {data/SAC/ALL/Slide.dat};
        \addplot [name path=HIGHQ, draw=none, forget plot] table [x={timestep}, y={highq}]{data/SAC/ALL/Slide.dat};
        \addplot [draw=none, fill=gray, opacity=0.2, color=color1, forget plot] fill between[of=LOWQ and HIGHQ];
        
        \addplot [color2] table [x={timestep}, y={med}] {data/SAC/no_HER/Slide.dat};
        \addplot [name path=LOWQ, draw=none, forget plot] table [x={timestep}, y={lowq}] {data/SAC/no_HER/Slide.dat};
        \addplot [name path=HIGHQ, draw=none, forget plot] table [x={timestep}, y={highq}]{data/SAC/no_HER/Slide.dat};
        \addplot [draw=none, fill=gray, opacity=0.2, color=color2, forget plot] fill between[of=LOWQ and HIGHQ];
        
        \addplot [color3] table [x={timestep}, y={med}] {data/SAC/no_DQT/Slide.dat};
        \addplot [name path=LOWQ, draw=none, forget plot] table [x={timestep}, y={lowq}] {data/SAC/no_DQT/Slide.dat};
        \addplot [name path=HIGHQ, draw=none, forget plot] table [x={timestep}, y={highq}]{data/SAC/no_DQT/Slide.dat};
        \addplot [draw=none, fill=gray, opacity=0.2, color=color3, forget plot] fill between[of=LOWQ and HIGHQ];
    \end{axis}
    
    \begin{axis}[
	    name=PickAndPlaceDDPG,
	    at={($(SlideDDPG.south)+(0, -1.0cm)$)},anchor=north,
        width=0.3\textwidth,
        grid=major,
        scaled x ticks = false,
        ylabel=success rate,
        xticklabels={,,},
        ymin=-0.1, ymax=1.1,
        ]
        \addplot [color1] table [x={timestep}, y={med}] {data/DDPG/ALL/PickAndPlace.dat};
        \addplot [name path=LOWQ, draw=none, forget plot] table [x={timestep}, y={lowq}] {data/DDPG/ALL/PickAndPlace.dat};
        \addplot [name path=HIGHQ, draw=none, forget plot] table [x={timestep}, y={highq}]{data/DDPG/ALL/PickAndPlace.dat};
        \addplot [draw=none, fill=gray, opacity=0.2, color=color1, forget plot] fill between[of=LOWQ and HIGHQ];
        
        \addplot [color2] table [x={timestep}, y={med}] {data/DDPG/no_HER/PickAndPlace.dat};
        \addplot [name path=LOWQ, draw=none, forget plot] table [x={timestep}, y={lowq}] {data/DDPG/no_HER/PickAndPlace.dat};
        \addplot [name path=HIGHQ, draw=none, forget plot] table [x={timestep}, y={highq}]{data/DDPG/no_HER/PickAndPlace.dat};
        \addplot [draw=none, fill=gray, opacity=0.2, color=color2, forget plot] fill between[of=LOWQ and HIGHQ];
    \end{axis}
    
    \draw (0:0) node[at={($(PickAndPlaceDDPG.west)+(-1.5cm,0)$)},  rotate=90]{\footnotesize\texttt{PandaPickAndPlace-v1}};
    
    \begin{axis}[
	    name=PickAndPlaceTD3,
	    at={($(PickAndPlaceDDPG.east)+(0.5cm,0)$)}, anchor=west,
        width=0.3\textwidth,
        grid=major,
        scaled x ticks = false,
        yticklabels={,,},
        xticklabels={,,},
        ymin=-0.1, ymax=1.1,
        ]
        \addplot [color1] table [x={timestep}, y={med}] {data/TD3/ALL/PickAndPlace.dat};
        \addplot [name path=LOWQ, draw=none, forget plot] table [x={timestep}, y={lowq}] {data/TD3/ALL/PickAndPlace.dat};
        \addplot [name path=HIGHQ, draw=none, forget plot] table [x={timestep}, y={highq}]{data/TD3/ALL/PickAndPlace.dat};
        \addplot [draw=none, fill=gray, opacity=0.2, color=color1, forget plot] fill between[of=LOWQ and HIGHQ];
        
        \addplot [color2] table [x={timestep}, y={med}] {data/TD3/no_HER/PickAndPlace.dat};
        \addplot [name path=LOWQ, draw=none, forget plot] table [x={timestep}, y={lowq}] {data/TD3/no_HER/PickAndPlace.dat};
        \addplot [name path=HIGHQ, draw=none, forget plot] table [x={timestep}, y={highq}]{data/TD3/no_HER/PickAndPlace.dat};
        \addplot [draw=none, fill=gray, opacity=0.2, color=color2, forget plot] fill between[of=LOWQ and HIGHQ];
        
        \addplot [color3] table [x={timestep}, y={med}] {data/TD3/no_DQT/PickAndPlace.dat};
        \addplot [name path=LOWQ, draw=none, forget plot] table [x={timestep}, y={lowq}] {data/TD3/no_DQT/PickAndPlace.dat};
        \addplot [name path=HIGHQ, draw=none, forget plot] table [x={timestep}, y={highq}]{data/TD3/no_DQT/PickAndPlace.dat};
        \addplot [draw=none, fill=gray, opacity=0.2, color=color3, forget plot] fill between[of=LOWQ and HIGHQ];
    \end{axis}
    
    \begin{axis}[
	    name=PickAndPlaceSAC,
	    at={($(PickAndPlaceTD3.east)+(0.5cm,0)$)},anchor=west,
	    width=0.3\textwidth,
        grid=major,
        scaled x ticks = false,
        yticklabels={,,},
        xticklabels={,,},
        ymin=-0.1, ymax=1.1,
        ]
        \addplot [color1] table [x={timestep}, y={med}] {data/SAC/ALL/PickAndPlace.dat};
        \addplot [name path=LOWQ, draw=none, forget plot] table [x={timestep}, y={lowq}] {data/SAC/ALL/PickAndPlace.dat};
        \addplot [name path=HIGHQ, draw=none, forget plot] table [x={timestep}, y={highq}]{data/SAC/ALL/PickAndPlace.dat};
        \addplot [draw=none, fill=gray, opacity=0.2, color=color1, forget plot] fill between[of=LOWQ and HIGHQ];
        
        \addplot [color2] table [x={timestep}, y={med}] {data/SAC/no_HER/PickAndPlace.dat};
        \addplot [name path=LOWQ, draw=none, forget plot] table [x={timestep}, y={lowq}] {data/SAC/no_HER/PickAndPlace.dat};
        \addplot [name path=HIGHQ, draw=none, forget plot] table [x={timestep}, y={highq}]{data/SAC/no_HER/PickAndPlace.dat};
        \addplot [draw=none, fill=gray, opacity=0.2, color=color2, forget plot] fill between[of=LOWQ and HIGHQ];
        
        \addplot [color3] table [x={timestep}, y={med}] {data/SAC/no_DQT/PickAndPlace.dat};
        \addplot [name path=LOWQ, draw=none, forget plot] table [x={timestep}, y={lowq}] {data/SAC/no_DQT/PickAndPlace.dat};
        \addplot [name path=HIGHQ, draw=none, forget plot] table [x={timestep}, y={highq}]{data/SAC/no_DQT/PickAndPlace.dat};
        \addplot [draw=none, fill=gray, opacity=0.2, color=color3, forget plot] fill between[of=LOWQ and HIGHQ];
    \end{axis}
    
    \begin{axis}[
	    name=StackDDPG,
	    at={($(PickAndPlaceDDPG.south)+(0, -1.0cm)$)},anchor=north,
        width=0.3\textwidth,
        grid=major,
        ylabel=success rate,
        xlabel=timesteps,
        ymin=-0.1, ymax=1.1,
        ]
        \addplot [color1] table [x={timestep}, y={med}] {data/DDPG/ALL/Stack.dat};
        \addplot [name path=LOWQ, draw=none, forget plot] table [x={timestep}, y={lowq}] {data/DDPG/ALL/Stack.dat};
        \addplot [name path=HIGHQ, draw=none, forget plot] table [x={timestep}, y={highq}]{data/DDPG/ALL/Stack.dat};
        \addplot [draw=none, fill=gray, opacity=0.2, color=color1, forget plot] fill between[of=LOWQ and HIGHQ];
        
        \addplot [color2] table [x={timestep}, y={med}] {data/DDPG/no_HER/Stack.dat};
        \addplot [name path=LOWQ, draw=none, forget plot] table [x={timestep}, y={lowq}] {data/DDPG/no_HER/Stack.dat};
        \addplot [name path=HIGHQ, draw=none, forget plot] table [x={timestep}, y={highq}]{data/DDPG/no_HER/Stack.dat};
        \addplot [draw=none, fill=gray, opacity=0.2, color=color2, forget plot] fill between[of=LOWQ and HIGHQ];
    \end{axis}
    
    \draw (0:0) node[at={($(StackDDPG.west)+(-1.5cm,0)$)},  rotate=90]{\footnotesize\texttt{PandaStack-v1}};
    
    \begin{axis}[
	    name=StackTD3,
	    at={($(StackDDPG.east)+(0.5cm,0)$)}, anchor=west,
        width=0.3\textwidth,
        grid=major,
        yticklabels={,,},
        ymin=-0.1, ymax=1.1,
        xlabel=timesteps,
        legend style={at=(0,-1.0cm), anchor=north}
        ]
        \addplot [color1] table [x={timestep}, y={med}] {data/TD3/ALL/Stack.dat};\label{pgfplots:no_ablation}
        \addplot [name path=LOWQ, draw=none, forget plot] table [x={timestep}, y={lowq}] {data/TD3/ALL/Stack.dat};
        \addplot [name path=HIGHQ, draw=none, forget plot] table [x={timestep}, y={highq}]{data/TD3/ALL/Stack.dat};
        \addplot [draw=none, fill=gray, opacity=0.2, color=color1, forget plot] fill between[of=LOWQ and HIGHQ];\label{pgfplots:_}
        
        \addplot [color2] table [x={timestep}, y={med}] {data/TD3/no_HER/Stack.dat}; \label{pgfplots:no_HER}
        \addplot [name path=LOWQ, draw=none, forget plot] table [x={timestep}, y={lowq}] {data/TD3/no_HER/Stack.dat};
        \addplot [name path=HIGHQ, draw=none, forget plot] table [x={timestep}, y={highq}]{data/TD3/no_HER/Stack.dat};
        \addplot [draw=none, fill=gray, opacity=0.2, color=color2, forget plot] fill between[of=LOWQ and HIGHQ];
        
        \addplot [color3] table [x={timestep}, y={med}] {data/TD3/no_DQT/Stack.dat};\label{pgfplots:no_DQT}
        \addplot [name path=LOWQ, draw=none, forget plot] table [x={timestep}, y={lowq}] {data/TD3/no_DQT/Stack.dat};
        \addplot [name path=HIGHQ, draw=none, forget plot] table [x={timestep}, y={highq}]{data/TD3/no_DQT/Stack.dat};
        \addplot [draw=none, fill=gray, opacity=0.2, color=color3, forget plot] fill between[of=LOWQ and HIGHQ];
        
        % \legend{A,B,C}
    \end{axis}
    
    \begin{axis}[
	    name=StackSAC,
	    at={($(StackTD3.east)+(0.5cm,0)$)},anchor=west,
	    width=0.3\textwidth,
        grid=major,
        xlabel=timesteps,
        yticklabels={,,},
        ymin=-0.1, ymax=1.1,
        ]
        \addplot [color1] table [x={timestep}, y={med}] {data/SAC/ALL/Stack.dat};
        \addplot [name path=LOWQ, draw=none, forget plot] table [x={timestep}, y={lowq}] {data/SAC/ALL/Stack.dat};
        \addplot [name path=HIGHQ, draw=none, forget plot] table [x={timestep}, y={highq}]{data/SAC/ALL/Stack.dat};
        \addplot [draw=none, fill=gray, opacity=0.2, color=color1, forget plot] fill between[of=LOWQ and HIGHQ];
        
        \addplot [color2] table [x={timestep}, y={med}] {data/SAC/no_HER/Stack.dat};
        \addplot [name path=LOWQ, draw=none, forget plot] table [x={timestep}, y={lowq}] {data/SAC/no_HER/Stack.dat};
        \addplot [name path=HIGHQ, draw=none, forget plot] table [x={timestep}, y={highq}]{data/SAC/no_HER/Stack.dat};
        \addplot [draw=none, fill=gray, opacity=0.2, color=color2, forget plot] fill between[of=LOWQ and HIGHQ];
        
        \addplot [color3] table [x={timestep}, y={med}] {data/SAC/no_DQT/Stack.dat};
        \addplot [name path=LOWQ, draw=none, forget plot] table [x={timestep}, y={lowq}] {data/SAC/no_DQT/Stack.dat};
        \addplot [name path=HIGHQ, draw=none, forget plot] table [x={timestep}, y={highq}]{data/SAC/no_DQT/Stack.dat};
        \addplot [draw=none, fill=gray, opacity=0.2, color=color3, forget plot] fill between[of=LOWQ and HIGHQ];
    \end{axis}

    \draw (0:0) node[at={($(StackTD3.south)+(0,-1.0cm)$)}, anchor=north]{
    \begin{tabular}{|clclcl|}
    \hline
    \ref{pgfplots:no_ablation} & Full &
    \ref{pgfplots:no_HER} & No HER &
    \ref{pgfplots:no_DQT} & No clipped double-$Q$ trick
    \\ \hline
    \end{tabular}};

\end{tikzpicture}
\caption{Ablation study for HER and for the clipped double-$Q$ trick. We repeat each experiment with 21 different random seeds. Median success rates are solid lines and interquartile range are shaded areas.}
\label{fig:learning_curves_ablations}
\end{figure}
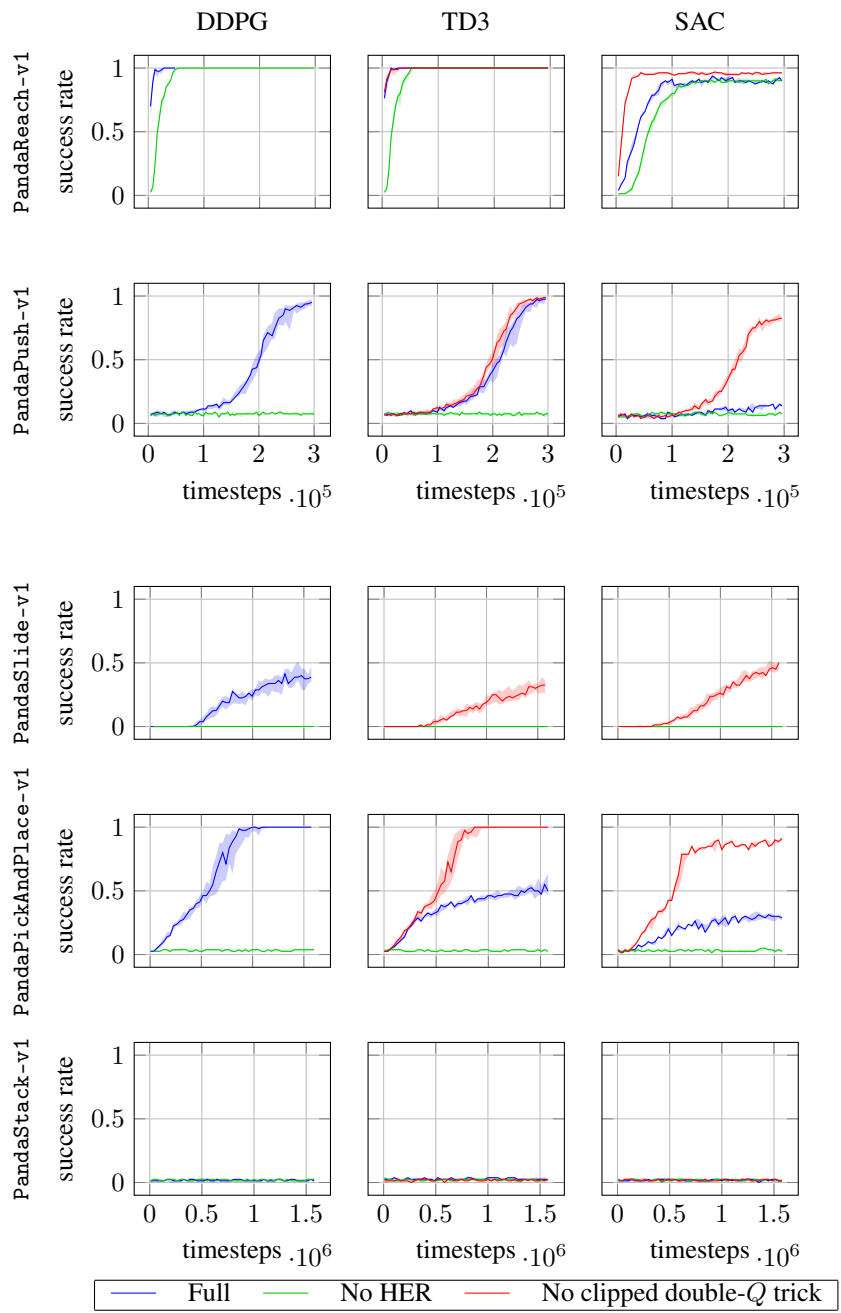

The results are presented in Figure \ref{fig:learning_curves_ablations}.

Although the clipped double-$Q$ trick allows in some environments to increase the learning performance (\texttt{Hopper-v2}, \texttt{Walker2d-v2}, \texttt{HalfCheetah-v2}, \texttt{Humanoid-v2}, \citep{haarnoja2018soft}), by limiting the value overestimation, its ablation leads here either to no effect, or to an increase of the results. It is possible that the scattering of the reward prevents overestimation of value. This trick would thus be counterproductive, by preventing the value from spreading properly. This intuition should be verified in future work.

\newpage
\section{Environments specifications}

\begin{table}[ht]
  \centering
\begin{tabular}{lcccccccc}
\toprule
     & \multicolumn{4}{c}{Observation} & \multicolumn{3}{c}{Action} \\ \cmidrule(r){2-8}
     Task & Gripper & Object(s)& Gripper& Size & Gripper & Gripper & Size\\
     & pose & pose & opening & & displacement & opening & \\ \hline
    Reach & \cmark & \xmark & \xmark & 6  & \cmark & \xmark & 3\\
    Push  & \cmark & \cmark & \xmark & 18 & \cmark & \xmark & 3\\
    Slide & \cmark & \cmark & \xmark & 18 & \cmark & \xmark & 3\\
    PickAndPlace & \cmark & \cmark & \cmark & 19 & \cmark & \cmark & 4\\
    Stack & \cmark & \cmark & \cmark & 31 & \cmark & \cmark & 4\\
    \bottomrule
\end{tabular}
\caption{Components of the observation and the action space.}
  \label{tab:observation space}
\end{table}

\section{Overview of the learned policies}
\label{appendix:policies}
\begin{figure}[ht]
    \centering
    \includegraphics[width=\textwidth]{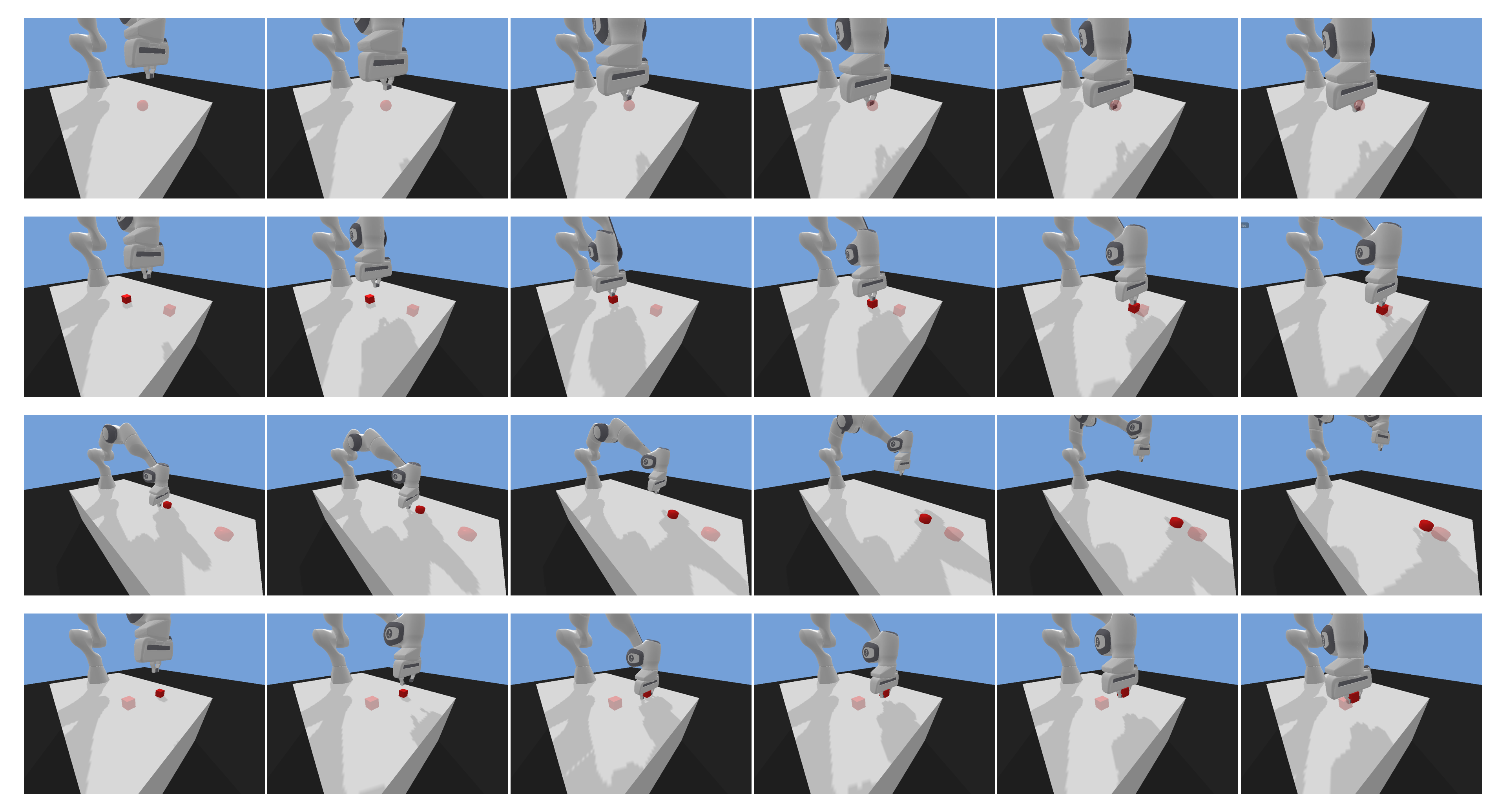}
    \caption{Overview of policies at the end of the training. Each line represents a task. From top to bottom: reach (one timestep between two successive images), push (two timesteps between two successive images), slide (four timesteps between two successive images) and pick \& place (two timesteps between two successive images).}
    \label{fig:policies}
\end{figure}

\end{document}